\documentclass[letterpaper, 10 pt, conference]{ieeeconf}  % Comment this line out if you need a4paper

\IEEEoverridecommandlockouts                              % This command is only needed if 
\usepackage{graphicx}
\graphicspath{ {images/} }
\usepackage{amsmath}
\usepackage{xcolor}
\usepackage{floatrow}
\usepackage{epstopdf}
\usepackage[colorlinks = true,
            linkcolor = black,
            urlcolor  = magenta,
            citecolor = black,
            anchorcolor = black]{hyperref}

\usepackage{algorithm,algpseudocode}
\usepackage[font=small]{caption}
\usepackage{accents}
\usepackage{amsfonts}
\usepackage{wrapfig}
\usepackage{booktabs}
\usepackage{balance}
\usepackage{scalerel}
\usepackage{subfigure}
\usepackage{url}
\usepackage{svg}

\newcommand{\norm}[1]{\left\lVert#1\right\rVert}

\newcommand{\zm}[1]{{\color{blue} \bf ZM: **#1**} }

\long\def\ignore#1{}
\long\def\lite{{\em LIT} }
\long\def\liteend{{\em LIT}}
\long\def\prolit{{\em ProLIT} }
\long\def\prolitend{{\em ProLIT}}
\title{\LARGE \bf
LIT: Light-field Inference of Transparency \\for Refractive Object Localization 
}

\author{Zheming Zhou\hspace{0.5cm} Xiaotong Chen\hspace{0.5cm} Odest Chadwicke Jenkins
\thanks{The authors are with the Department of Electrical Engineering and Computer Science, Robotics Institute, University of Michigan, Ann Arbor, MI, USA, 48109-2121  {\tt\small [zhezhou|cxt|ocj]@umich.edu}}
}

\begin{document}

\maketitle
\thispagestyle{empty}
\pagestyle{empty}
\graphicspath{{images/}}

%%%%%%%%%%%%%%%%%%%%%%%%%%%%%%%%%%%%%%%%%%%%%%%%%%%%%%%%%%%%%%%%%%%%%%%%%%%%%%%%
\begin{abstract}

Translucency is prevalent in everyday scenes. As such, perception of transparent objects is essential for robots to perform manipulation. Compared with texture-rich or texture-less Lambertian objects, transparency induces significant uncertainty on object appearances. Ambiguity can be due to changes in lighting, viewpoint, and backgrounds, each of which brings challenges to existing object pose estimation algorithms. In this work, we propose \liteend, a two-stage method for transparent object pose estimation using light-field sensing and photorealistic rendering. \lite employs multiple filters specific to light-field imagery in deep networks to capture transparent material properties, with robust depth and pose estimators based on generative sampling. Along with the \lite algorithm, we introduce the light-field transparent object dataset \prolit for the tasks of recognition, localization and pose estimation. With respect to this \prolit dataset, we demonstrate that \lite can outperform both state-of-the-art end-to-end pose estimation methods and a generative pose estimator on transparent objects. The link of supplementary material can be found at: \href{https://sites.google.com/umich.edu/prolit}{https://sites.google.com/umich.edu/prolit}

\end{abstract}

\section{INTRODUCTION}

% see intro of DOPE, PoseCNN as reference

% lightfield sfm: https://ieeexplore.ieee.org/stamp/stamp.jsp?tp=&arnumber=8556460

% Taking about the 
% 1. difference between 4D light-field pose estimation vs. 2D image based pose estimation.
% 2. difference between transparent object pose estimation vs. texture-rich  vs. texture-less with Lambertain surface. 

Recognizing and localizing objects has a wide range of applications in robotics, and remains a very challenging problem. The challenge comes from the variety of objects in the real world and the continuous high dimension spaces of object poses. The diversity of object materials also induces strong uncertainty and noise for sensor observations. Existing works and datasets~\cite{sui2017sum, wang2019densefusion, tremblay2018deep} cover a variety of texture-rich objects with distinguishable features between different types of objects. Several other works~\cite{sundermeyer2018implicit, li2018deepim} cover textureless objects with Lambertian surfaces, where robot sensors can still perceive rich depth information. 
% However, transparent objects are also prevalent in the real world.  
However, many of these assumptions for objects with Lambertian surface properties are ill-posed for transparent objects.
%in robot perceptions are not satisfied for them.

The challenges imposed by transparency are multidimensional. First, non-Lambertian surface texture is highly reliant on the environment lighting and background appearance. Specifically, transparent surfaces will produce specularity from environmental lighting and project distorted background texture on their surfaces due to refraction. Second, transparent object depth information cannot be correctly captured by RGB-D sensors, which are commonly used by current object recognition and localization methods. This limitation imposes difficulties in collecting transparent object pose data using current labeling tools~\cite{marion2018label}. As a result, transparent objects remain effectively invisible to robots using the sensors.
\begin{figure}[t!]
\centering
\includegraphics[width=\columnwidth]{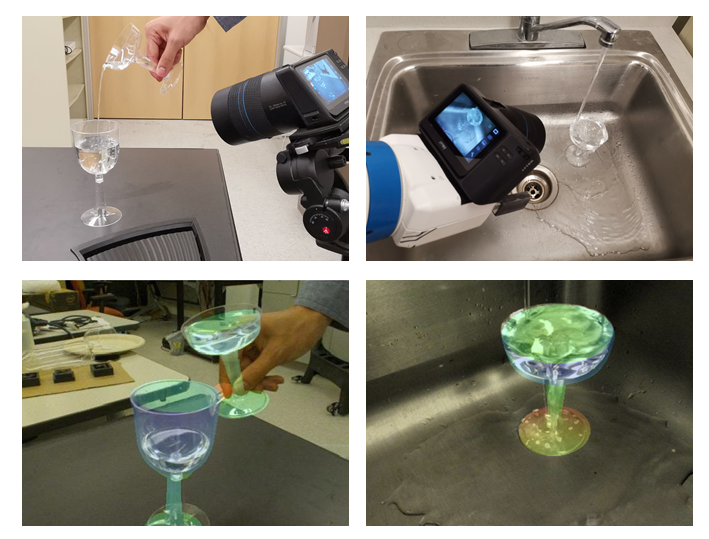}
\caption{Demonstration of our \lite pipeline. (Top row) Lytro Illum camera is mounted on the tripod and robot arm to capture the transparent objects in challenging environments. (Bottom row) final estimated poses are overlapped to the center view of the observed light-field image.}
\end{figure}

Recently, several works~\cite{zhou2018plenoptic, oberlintime} showed promising results using light-field (or plenoptic) photography in perceiving transparent objects. For example, Zhou \textit{et al.}~\cite{zhou2019glassloc} generated grasp poses for transparent objects by classifying local patch features in a \textit{Depth Likelihood Volume (DLV)} plenoptic descriptor. However, capturing and labeling over light-field images is time-consuming and computationally costly. Synthetic data is an alternative for image generation and has shown encouraging results in object recognition and localization. Georgakis \textit{et al.}~\cite{georgakis2017synthesizing} rendered photorealistic images by projecting the object texture model on the real background for training object detectors. Tremblay \textit{et al.}~\cite{tremblay2018deep} proposed DOPE as an end-to-end pose estimator using domain randomization and photorealistic rendering~\cite{to2018ndds}. We similarly address the problem of transparency using photorealistic rendering and light-field perception. 

\begin{figure*}[thpb]
\vspace{+1em}
   \centering
      \includegraphics[width=\textwidth]{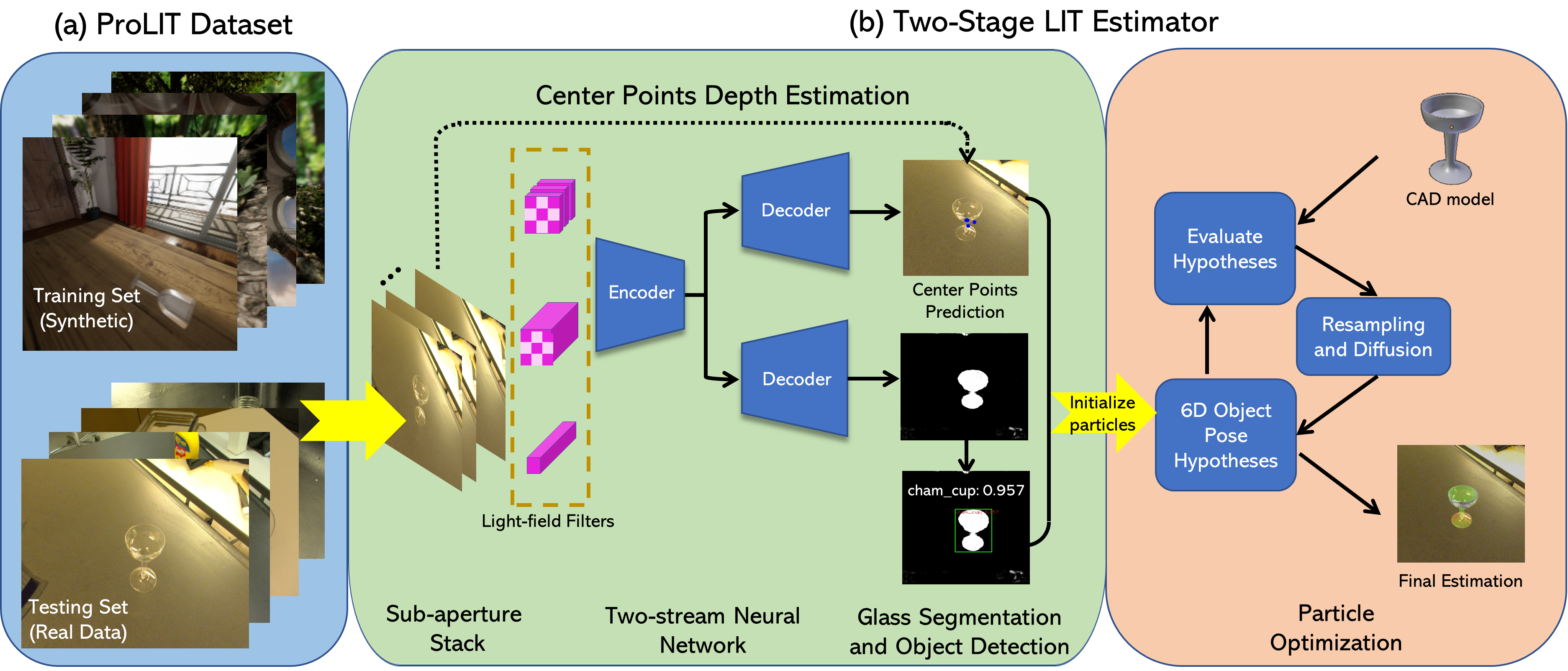}
      \caption{An overview of the \lite framework with the \prolit dataset. (a) \prolit contains 75,000 synthetic light-field images in training set and 300 real images with 442 object instances in testing set. (b) \lite estimator is a two-stage pipeline. The first stage takes light-field images as input and outputs transparent material segmentation and object center point prediction. The segmentation results are passed through a detection network to obtain object labels. In the second stage, for each predicted center point, we predict point depth likelihood by local depth estimation using Depth Likelihood Volume. The particle optimization samples over center points and converge to the pose that best matches the segmentation results.}
      \label{fig:pipeline}
\end{figure*}

In this paper, we propose \lite as a generative-discriminative method for recognition and pose estimation for transparent objects. 
%More specifically, our contributions are two-fold: \chad{this sentence is unnecessary}
Within \liteend, we introduce 3D convolutional light-field filters as the first layer of our neural network. This neural network is trained purely with synthetic data from a customized light-field rendering system for virtual environments. At run time, the output of this trained neural network is used as input to a generative inference. The pose estimates resulting from this inference are then used to perform grasping and manipulation tasks.
% We leverage network outputs with generative inference to achieve 6D pose estimation. 
We introduce the ProgressLIT light-field dataset (\prolitend) for the task of transparent objects recognition, segmentation, and pose estimation. The \prolit dataset contains 75,000 synthetic light-field images and 300 real images from Lytro Illum light-field camera labeled with segmentation and 6D object poses.
We show the efficacy of \lite with respect to state-of-the-art end-to-end methods and a generative method on our proposed \prolit transparent object dataset. We additionally present a demonstration of using \lite for a purposeful manipulation task of building a champagne tower in a sparsely textured environment.

\ignore{
a two-stage transparent object 6D pose estimation algorithm with the first stage trained with pure light-field synthetic data. 
To the best of authors' knowledge, this is the first time a complete pipeline is proposed on 6D pose estimation focused on transparent objects. We also propose a dataset on transparent objects segmentation and 6D pose estimation, with evaluation of \lite compared with state-of-the-art deep learning methods on object pose estimation and general approach on transparent objects estimation. Experiment results show that \lite outperforms baseline methods ...[\zm{More insight and contributions}]}

% and gives usable poses for robot manipulation tasks through the combinations of light-field inputs, discriminative power of deep learning and robustness of probabilistic inference.

\section{RELATED WORK}
\subsection{Pose Estimation for Robot Manipulation}

6D pose estimation remains a central problem in robot perception for manipulation in recent years. Deep learning methods have been a prevalent approach to perform accurate and fast inference for this problem. Xiang \textit{et al.}~\cite{xiang2017posecnn} proposed PoseCNN to recognize and estimate objects and their 6D poses by decoupling translation and rotation separately in a neural network structure. Other end-to-end method methods have explored using synthetic data in training~\cite{tremblay2018deep,josifovski2018object}, pixel-wise voting over keypoints~\cite{hu2019segmentation,peng2019pvnet}, and residual networks to iteratively refine object poses~\cite{li2018deepim,wang2019densefusion}. Hybrid (or generative-discriminative) methods can achieve better performance by using deep networks to give hypotheses of object poses followed by a second stage of refinement. To get the final pose estimates, a variety of methods have been proposed for the second stage, including probabilistic generative inference~\cite{sui2017sum,chen2019grip}, template matching~\cite{park2019multi}, and point cloud registration~\cite{sundermeyer2018implicit,mitash2018robust}.
 
Most deep learning methods for pose estimation are focused on texture-rich objects or those with texture-less but Lambertian surfaces~\cite{park2019multi,sundermeyer2018implicit}. Transparent objects bring challenges in two main aspects, where there is: 1) no reliable depth information, and 2) no distinguishable environment-independent color textures. Prior works~\cite{lysenkov2013recognition,lysenkov2013pose} have used invalid readings from depth camera to extract object contours for pose estimation. However, these methods rely on the Lambertain reflections of the background surface to establish reliable contour of transparent objects. We take inspiration from these ideas for perception from light-field observations in two ways. First, a decent detection or segmentation intermediate result plays an important role in restricting the search area of the 6D object pose. Further, a deep network trained on a large, elaborately designed synthetic dataset can reach similar performance with those trained on real world data.

\subsection{Light-field Perception for Transparency}

The foundation of light-field image rendering was first introduced by Levoy and Hanrahan~\cite{levoy1996light} for the purpose of sampling new views from existing images. Since the seminal work, light-field cameras have shown advancement in performing visual tasks in challenging environments with transparency and translucency. Maeno \textit{et al.}~\cite{maeno2013light} proposed the light-field distortion features from epipolar images for recognizing transparent objects. Recent work by Tsai \textit{et al.}~\cite{tsai2018distinguishing} further explored the light-field features to distinguish transparent and Lambertian materials. The result showed that the distortion features in the epipolar images can be used to distinguish materials with different refraction properties. Apart from refraction, specular reflection is another unique property carried by transparent materials. Tao \textit{et al.}~\cite{tao2015depth} investigated the line consistency in the light-field images with a dichromatic reflection model that removes the specularity from the images. Alperovich \textit{et al.}~\cite{alperovich2018light} proposed fully convolutional networks to separate specularity in light-field images. In robotics, Zhou \textit{et al.}~\cite{zhou2018plenoptic, zhou2019glassloc} created a plenoptic descriptor called DLV to model the depth uncertainty in a layered translucent environment. Based on this DLV, the object poses and grasp poses for robot manipulation  are estimated using generative inference. Our proposed \lite method is built on these ideas above and leverages the power of discriminative and generative methods with data generation using photorealistic rendering. 

\section{LIT ESTIMATOR} \label{sect:method}
% \zm{Change the problem formulation to add clear Input-Output variables and definitions}

Given an input light-field image $L$, the objective of \lite estimator is to infer the objects label $l$ and their poses $q$ in $SE(3)$. The pose $q$ represents the transformation from object local coordinate frame to the camera coordinate frame. For a light-field image $L$ with spatial resolution $H_s\times W_s$ and angular resolution $H_a\times W_a$, we assume the camera coordinate frame overlaps with the center view image's coordinate frame. The object pose $q$ is defined in center view and parameterized into 3D translation and 3D orientation in quaternion.

\subsection{LIT Pipeline}
The two-stage \lite pipeline is shown in Figure~\ref{fig:pipeline}.
The first stage consists of a two-stream neural network that outputs pixel-wise image segmentation and 2D object center point locations. This output is followed by a detection network that classifies object labels $l$ and clusters the corresponding center points. A light-field based object depth estimator gives object center depth distributions. The second-stage is a particle optimization initialized based on network and depth estimates, that converges to the final 6D poses.

There are several insights incorporated in the pipeline design. First, the segmentation decoder branch in the first neural network performs transparent material segmentation rather than object-class or instance segmentation. This distinction means it only decides whether a pixel belongs to a transparent material or not.
% The object classification problem is further settled in the following detection network.
The rationale for this classification is that pixel values within transparent object areas highly depend on the background and material property, rather than object types. Thus, it is difficult for a single network to distinguish different objects from raw pixel values. In addition, the center point estimation branch does not regress multiple keypoints which is common in texture-rich object pose estimation networks \cite{hu2019segmentation, peng2019pvnet}. The further rationale is that transparent objects lack features that are independent to object poses and environmental changes, such as background and lighting. In our work, we only predict the 2D object center point location.
% In other words, the same point on the object may have various appearances. In our work, we find the networks perform worse in end-to-end object-wise segmentation, and they fail in differentiating 3D bounding keypoints except the center point.

% The pipeline Our goal is to estimate 6D poses (including 3D location and 3D orientation) of transparent objects given a light-field image. The input image first passes through three customized filters for extracting geometric features from the difference between sub-aperture images. The extracted light-field feature vectors then pass an encoder-decoder architecture where the two decoding branches give pixel-wise semantic segmentation and object keypoints estimation. Object label classification and keypoint clustering are done in a detection network. Object depth is obtained from a light-field image based estimator. Finally, a sampling-based particle optimization process outputs the 6D pose estimation.

\subsection{Network Architecture}
\begin{figure}[t!]
\includegraphics[width=0.95\columnwidth]{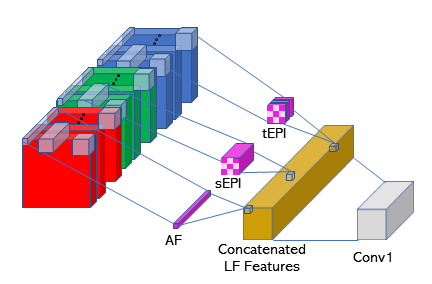}
\caption{Illustration of three light-field filters. Angular filter (AF) has dimension $1\times1\times(H_a \times W_a)$ to capture features in angular pixels. sEPI and tEPI filters have sizes of $n\times n\times W_a$ and $n\times n\times H_a$ respectively, here $n$ refers to kernel size. tEPI also has a dilation $W_a$. All features will be concatenated together after passing filters.}
\end{figure}
As shown in Figure~\ref{fig:pipeline}, the input light-field image is first decomposed into sub-aperture image stacks. This structure gives a 3D matrix with size $H_s \times W_s \times (H_a \times W_a)$ replicated for each of the R, G, B channels. The stacks are then going through three light-field filters: angular filter~\cite{wang20164d}, 3D sEPI filter, and 3D tEPI filter. 

\begin{itemize}
    \item \textbf{Angular Filter}. The angular filter aims to capture the reflection property of 3D surface points in the direction space of light ray. For instance, a non-Lambertian surface will establish different colors in a single angular patch while it will be nearly identical for a Lambertian surface. 
    % The output of the angular filter can be expressed as:
    % \begin{equation}
    % l_{a}^j(x,y) =  g(\sum_{i=r,g,b}\sum_{s,t}w^j_i(s,t)L(x,y,s,t)) 
    % \end{equation}
    % where $L(x,y,s,t)$ is the 4D light field function. $j$ is the number  of filters. \cxt{index of feature} $g(\cdot)$ is the ReLU function and $l^j(x,y)$ is light-field intermediate layer.
    The angular filter can be expressed as an operation over each pixel $(x, y)$ in spatial space (for the $j$th filter):
    \begin{equation} g(\sum_{s,t}  w^j_i(s,t)L_i(x,y,(s,t)))
    \end{equation}
    where $g(\cdot)$ is the activation function, $s$ and $t$ are the angular indices, $w_i^j$ is the weight in the angular filter, $i\in\{r,g,b\}$ is the color channel, and $L_i(x,y,(s,t))$ is the 4D light-field function.
    
    \item \textbf{3D EPI Filters}. Transparent surfaces will produce distortion features because of refraction. In the epipolar image plane, it will produce polynomial curve patterns which can be distinguished from the background texture without distortion. To capture distortion features, we propose the epipolar filters using 3D convolution layers along the two angular dimensions $s$ and $t$ respectively. The 3D EPI filters can be expressed as:
    \begin{equation}
    \begin{split} g(\sum_{u,v,s}\tilde{w}^j_i(u,v,s)L_i(x+u,y+v,(s,t)))\\
    g(\sum_{u,v,t}\hat{w}^j_i(u,v,t)L_i(x+u,y+v,(s,t)))
    \end{split}
    \end{equation}
    where $(u,v)$ is the index of convolution kernel in spatial space, $\tilde{w}$, $\hat{w}$ are weights in sEPI and tEPI filters, and we assume the input and output have the same dimension in spatial space by proper paddings.
\end{itemize}

Passing through the three customized filters, the embedded features of light-field images are concatenated. The result goes into an encoder-decoder structure with two branches for image segmentation and object center point regression. The output of the segmentation branch is a pixel-wise segmentation of the center view image. Each center view pixel is then predicted to be on a transparent surface, in the background, or on the boundary between a transparent object and background in the image. The output of the center point branch are the 2D pixel offsets from each pixel to their estimated center position on the image, as well as a pixel-wise confidence values.

The loss in segmentation branch $\mathcal{L}_{seg}$ is defined as the cross-entropy loss normalized by class pixel probabilities \cite{lin2017focal}. The loss of center point regression is mainly following design in \cite{hu2019segmentation}, although we only regress the center point positions. The learning goal for each pixel $p$ inside the segmentation area $\mathcal{M}$ is to regress the offset $h_p$ from its location $c_p$ to the object center $g_p$ on 2D image. In this way, the loss $\mathcal{L}_{pos}$ is expressed as:
\begin{equation}
    \mathcal{L}_{pos} = \sum_{p \in \mathcal{M}} \norm{g_p - (c_p + h_p)}_1
\end{equation}
where $\norm{\cdot}_1$ denotes $L^1$ loss. Each pixel's estimation is associated with a confidence value $b_p$, and the confidence loss $\mathcal{L}_{conf}$ is defined as:
\begin{equation}
    \mathcal{L}_{conf} = \sum_{p \in \mathcal{M}} \norm{b_p - \exp{(-\tau\norm{g_p - (c_p + h_p)}_2})}_1
\end{equation}
where $\tau$ is a modulating factor and $\norm{\cdot}_2$ denotes $L^2$ loss. The overall loss $\mathcal{L}$ is calculated as:
\begin{equation}
    \mathcal{L} = \alpha \mathcal{L}_{seg} + \beta \mathcal{L}_{pos} + \gamma \mathcal{L}_{conf}
\end{equation}
where $\alpha, \beta, \gamma$ modulates the importance of segmentation, regression and regression confidence respectively. In practice, we select $\alpha=1, \beta=8, \gamma=2$ from initial experimentation.

An object detection network is appended to differentiate object types based on geometry shapes from segmentation results. Specifically, the network takes the result of segmentation decoder branch as input and gives bounding boxes with object labels. Detected bounding boxes also play the role of clustering object center points. The overall output of the first stage is a set of bounding boxes, each with an object label and a set of object center points, which serves as the initial distribution of object center locations for the next stage.

Directly regressing the depth of center points without depth observation
is difficult for neural networks. Instead, we deploy a DLV plenoptic descriptor~\cite{zhou2018plenoptic} to describe the depth of a single pixel as a likelihood function rather than a deterministic value. The advantage of using a DLV is that depth likelihood can be naturally leveraged into generative inference framework in a sample initialization step. The likelihood $D(x_c,y_c,d)$ of a given center point located at $(x_c,y_c)$ in center view image plane $I_{c}$ can be calculated as:
\begin{equation}
    \label{eqn:dlv}
D(x_c,y_c,d) =
\frac{1}{N}\sum_{\scaleto{a\in A\backslash I_{c}}{6pt}}T_{a,d}(x_c,y_c)
\end{equation}
where $A$ is a set of sub-aperture views, $T_{a,d}(x_c,y_c)$ is the function to calculate the color intensity and gradient cost of pixel $(x_c,y_c)$ on a specific depth $d$. $\frac{1}{N}$ is a normalization term that maps cost to likelihood. Detailed implementation can be referred in~\cite{zhou2018plenoptic, zhou2019glassloc}.

\subsection{Particle Optimization}

The second stage of pipeline estimates the 6D pose of transparent objects in a sampling-based iterative likelihood reweighting process~\cite{mckenna2007tracking}. Object pose samples are initialized based on the center point locations from the first stage. During the iterations, rendered samples are projected to 2D image and their likelihoods are calculated as the similarity between the projected rendered samples and segmentation results.

\subsubsection{Sample Initialization}

Each sample is a hypothesis of object 6D pose. Its 3D location can be derived from 2D image coordinate $(u,v)$, depth $d$ and camera parameters. In this way, the probability distribution of 3D center point locations is formed by leveraging center point candidates and depth likelihood volume results:
\begin{equation}
% \left\{
\begin{split}
   & u = c_x + f_x\dfrac{x}{z},\quad
    v = c_y + f_y\dfrac{y}{z},\quad
    d = z\\
   & p(X=x, Y=y, Z=z) = b(u, v) D(u, v, d)
\end{split}
% \right.
\end{equation}
where $b$ is the confidence value of object center point estimation from neural network, $f_x, f_y, c_x, c_y$ are camera intrinsic parameters, and $D$ is likelihood from DLV in Equation~\eqref{eqn:dlv}. We perform importance sampling over this distribution to initialize the pose sample locations. The initial orientations of samples are randomly selected in $SO(3)$ space.

\begin{figure*}
\centering
    \subfigure[Training Set]{
        \centering
        \includegraphics[width=0.32\textwidth]{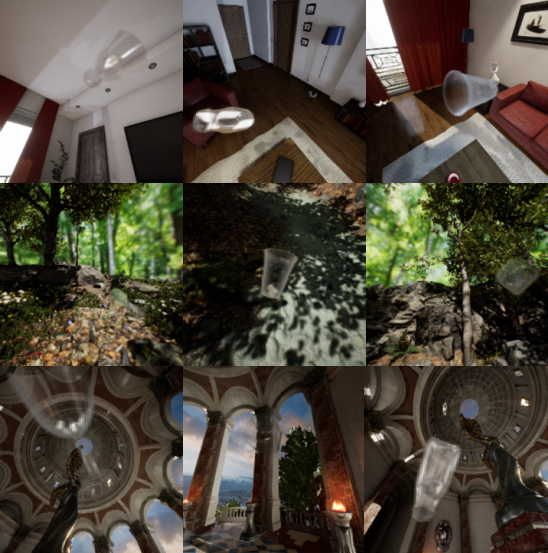}}
    \subfigure[Testing Set]{
        \centering
        \includegraphics[width=0.32\textwidth]{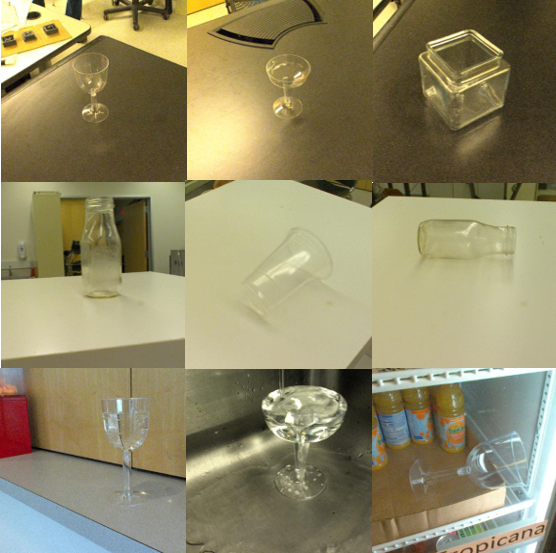}}
    \subfigure[Result]{
        \centering
        \includegraphics[width=0.32\textwidth]{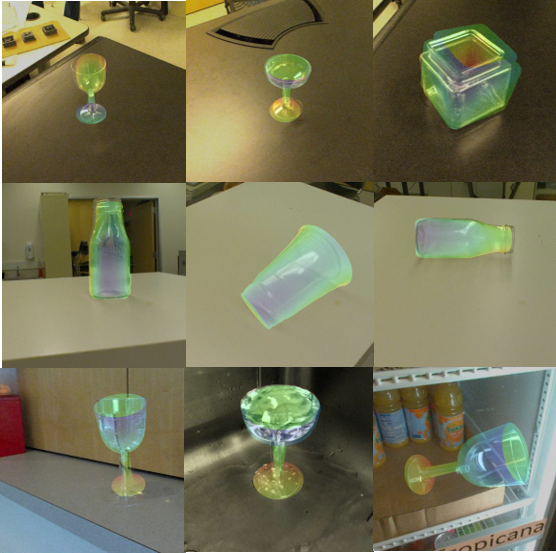}}
    \caption{(Left) example synthetic light-field images rendered in three different environments. (Middle) example test images in different backgrounds and different pose configurations. (Right) results visualization by overlaying estimated poses to the original test images.}
\label{fig:dataset}
\end{figure*}

\subsubsection{Likelihood Function}

The probability of each sample during iterations is calculated using the likelihood function, represented as the similarity between the projected rendered object point cloud and segmentation results from neural network. Specifically, the object points in its local frame are transformed by the sample pose and then projected to 2D image plane. The likelihood function is composed of intersection over union scores of projected rendered point clouds and segmentation masks on transparent material and its boundary:
\begin{equation}
    weight = \eta \dfrac{\vert S_{pcd} \cap S_{seg}\vert}{\vert S_{pcd} \cup S_{seg} \vert} + (1 - \eta) \dfrac{\vert \partial S_{pcd} \cap \partial S_{seg}\vert}{\vert \partial S_{pcd} \cup \partial S_{seg}\vert}
\end{equation}
where $S_{pcd}$ is the silhouette of projected rendered point cloud, $S_{seg}$ is the pixels segmented as transparent materials, $\partial S_{pcd}$ and $\partial S_{seg}$ are the sets of boundary pixels of $S_{pcd}$ and $S_{seg}$ respectively. $\eta$ is set to modulate importance of boundaries.

% \begin{figure}%
% \centering
%     \subfigure[Sample1]{%
%         \label{fig:first}%
%         \includegraphics[width=0.4\textwidth]{dataset.png}}%
%     \qquad
%     \subfigure[Sample2]{%
%         \label{fig:second}%
%         \includegraphics[width=0.4\textwidth]{dataset.png}}%
% \caption{sample}
% \end{figure}

\subsubsection{Update Process}

We follow the procedure of iterative likelihood reweighting to produce pose estimations. The initialized samples are assigned the same weights. Then the process of calculating likelihood values, resampling based on weights, and sample diffusion is repeated in every iteration. During diffusion step, each pose sample is randomly diffused in $SE(3)$ space in translation and rotation with Gaussian noise. The algorithm terminates when the maximum sample weight reaches a threshold, or the iteration number reaches the limit.
\section{PROLIT LIGHT-FIELD DATASET}
We propose the \prolit light-field image dataset for the task of transparent object recognition, segmentation, and 6D pose estimation. This dataset contains a total of 75,000 synthetic images and 300 real-world images with 442 object instances, each labeled with pixel-wise semantic segmentation and 6D object poses. Figure~\ref{fig:dataset} shows examples of synthetic images, real-world images and estimation results from \liteend. There are 5 instances of objects included in the dataset: wine cup, tall cup, glass jar, champagne cup, starbucks bottle with different geometric shapes. The images are captured using a Lytro Illum camera which is calibrated by the toolbox described in~\cite{bok2017geometric}. The spatial resolution of the calibrated image is $383\times552$, and the angular resolution is $5\times5$ (extracted from $9\times9$ sub-aperature images with stride 2). The object poses in testing data are labeled by reprojecting objects directly into the center view image and matching with observations. 

% The captured real data is treated as the testing set for \lite algorithm. For training the two-stream network of \lite pipeline, we use rendered light-field images which are also included in the dataset with generation tools

The light-field rendering pipeline is built on NDDS~\cite{to2018ndds} synthetic data generation plugin in Unreal Engine 4 (UE4). The created virtual light-field capturer has an angular resolution $5\times5$ and spatial resolution $224\times224$. The baseline between the adjacent virtual camera is set to $0.1$cm. We generate data in three UE4 world environments: room, temple, and forest. In each environment, we highly randomized the lighting conditions including color, direction, and intensity. The target objects are rendered using the transparent material. Objects move in two ways in the environment: flying in the air with random translation and rotation, or falling freely with collision and gravity enabled. When the objects move, the virtual light-field capturer will track and look at them with arbitrary azimuths and elevations. Ray tracing is enabled when capturing images.

\section{EXPERIMENTS}
% Input light-field images have spatial resolution $224 \times 224$ and angular resolution $5 \times 5$. 
We choose 64 light-field filters as the first feature extraction layer. The \lite network uses VGG16~\cite{simonyan2014very} as backbone architecture and initialized with pre-trained model on ImageNet~\cite{imagenet_cvpr09}. The segmentation branch outputs pixel-wise labels from over three classes: background, transparent, boundary. The center points prediction branch outputs pixel-wise offset for each segmented pixels. 
The detection network is a Faster R-CNN network~\cite{ren2015faster} with VGG16 backbone. The input to the network is the binary masks of transparent object segmentation and the output is bounding boxes with object labels. 

\subsection{Evaluation of Light-field Filters on Image Segmentation}

Segmentation is taken as the optimization target in our second stage which is critical to \lite pipeline. We first compare with two baseline methods to show the advantage of using light-field images with three light-field specific filters. One baseline takes input of 2D center view image, which passes through the same neural network structure as \lite except for light-field filters, the other is an ablation study with only the angular filter. All three networks are trained on the synthetic dataset containing 75,000 images. Table~\ref{tab:seg_comp} shows segmentation accuracy results, where \lite achieves better performance than baseline methods in all metrics. Through the comparison with single RGB input, we show that lighting direction information captured inside light-field images helps distinguish transparent pixels from the background.
% \lite outperforms the baseline method with single RGB input, which indicates that light-field images' capacity in capturing the direction of light can help in transparent material segmentation. 
Through the comparison with only an angular filter, \lite also achieves higher accuracy, showing that both angular features and EPI features are important in contributing to segmenting transparent objects.

\begin{table}
    \centering
    \small
    \begin{tabular}{c|c|c|c|c|c}
    \hline
    Method & gAcc & mAcc & mIoU & wIoU & mBFS\\
    \hline\hline
        2D & 0.871 & 0.500 & 0.228 & 0.397 & 0.140\\
    \hline
        AF only& 0.917 & 0.501 & 0.318 & 0.582 & 0.197\\
    \hline
        \textbf{\lite} & \textbf{0.954} & \textbf{0.520} & \textbf{0.455} & \textbf{0.854} & \textbf{0.390}\\
    \hline
    \end{tabular}
    \caption{Comparison of \lite and baseline methods on transparent material segmentation. The performance is quantified through global accuracy (gAcc), mean of class accuracy (mAcc), mean of Intersection over Union (mIoU), weighted IoU (wIoU), and mean BF (Boundary F1) contour matching score (mBFS). The definitions are detailed in~\cite{csurka2013good}. `AF only' here refers to the baseline method with only angular filters.}
    \label{tab:seg_comp}
\end{table}

\subsection{Evaluation of Pose Estimation}
We compare the 6D pose estimation results of \lite against a state-of-the-art general-purpose end-to-end object pose estimator, DOPE~\cite{tremblay2018deep}, a state-of-the-art textureless object pose estimator, Augmented Autoencoder (AAE)~\cite{sundermeyer2018implicit}, and a generative light-field based transparent object pose estimation method, PMCL~\cite{zhou2018plenoptic}. 
% Since DOPE outperforms PoseCNN~\cite{xiang2017posecnn} with pure synthetic data which itself outperforms many other single-shot pose estimation networks and AAE is a leading work in pose estimation of textureless objects, which are very similar to transparent objects, the comparison between \lite with DOPE and AAE can show our capability on transparent object pose estimation. 
% On the other hand, as depth is not available from transparent objects, we do not compare with deep neural network estimators using RGB-Depth data.

\begin{figure}[h]
    \centering
    \includegraphics[width=\columnwidth]{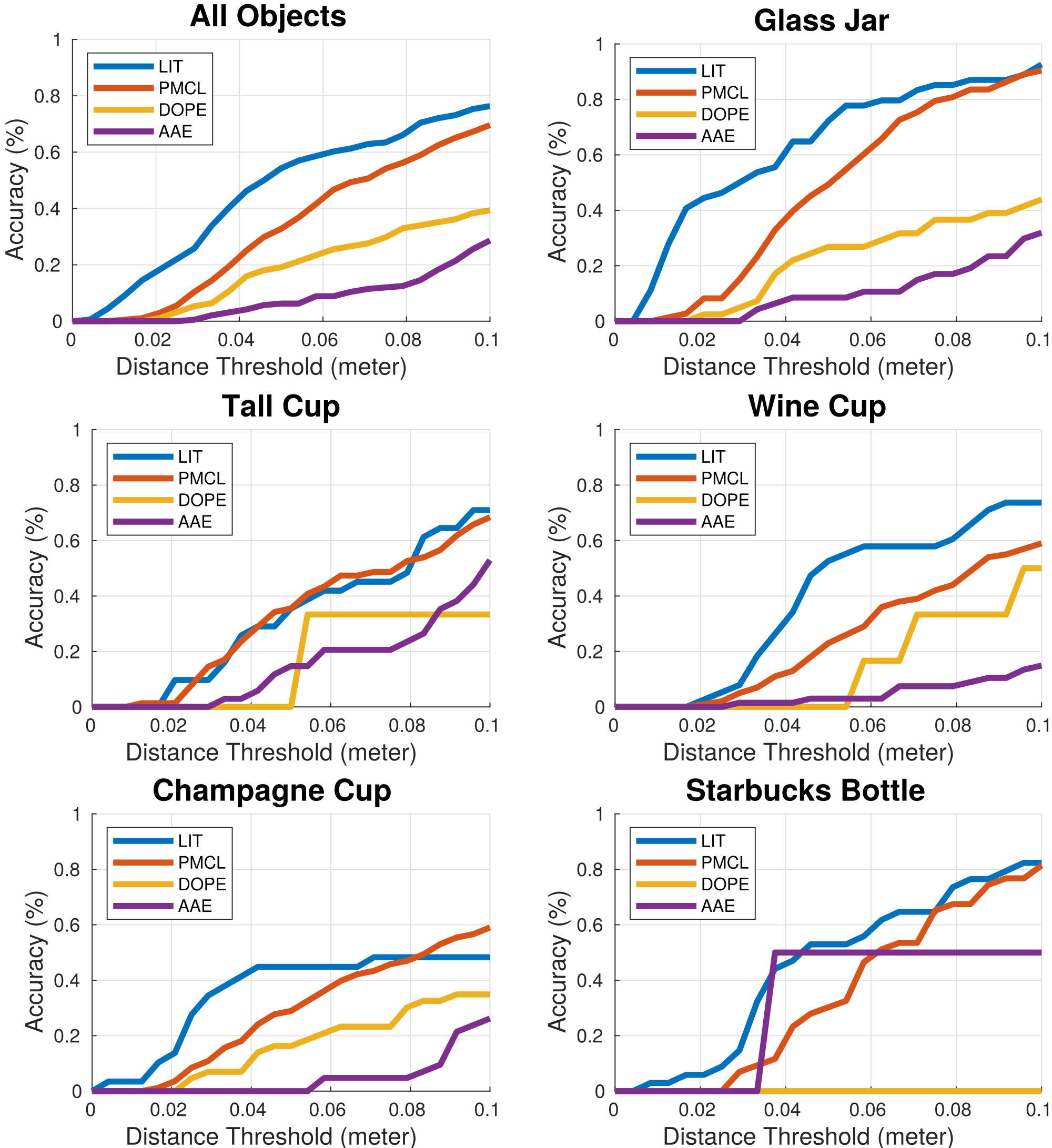}
    \caption{Comparison of 6D pose estimation results with respect to ADD-S and Accuracy Under Curve metric.}
    \label{fig:pose_comp}
\end{figure}

For the fair comparison with DOPE and AAE, we make both methods compatible with light-field inputs. We add the three light-field filters in Section~\ref{sect:method} before the first encoder layer of DOPE network as well as AAE encoder network. We adopt Faster R-CNN network as the first stage object detector for AAE. All of the methods are trained with 75,000 synthetic images for 5 objects. In the second stage of \lite pipeline, we diffuse the particles with Gaussian noise $\mathcal{N}$(0, 0.08) in translation and $\mathcal{N}$(0, 0.4) in orientation. PMCL is a generative method which requires object labels and 3D search space. We initialize PMCL with ground truth object labels and a search volume with size $40\times40\times40$ cm$^3$ around the ground truth object locations. The convergence threshold of particle weights is set to 0.7. We use ADD-S metric~\cite{xiang2017posecnn} to evaluate the pose results of symmetric objects. We then show the accuracy curves in Figure~\ref{fig:pose_comp} with a distance threshold of 0.1m. The Area Under accuracy-threshold Curve (AUC) and algorithm computation time per object are shown in Table~\ref{tab:pose_comp}.

From the result plots, we find that \lite performs much better than DOPE and AAE, and better than PMCL. For DOPE, we believe directly regress the eight 3D bounding box vertices and their relations is not an optimal strategy for transparent objects. First, DOPE's object recognition is embedded in the network but the transparent object's texture is not informative to distinguish different objects. Secondly, the eight vertices of 3D bounding boxes are ambiguous for networks to learn the features because of the object symmetry and lack of distinguishable features for transparent objects. For AAE, it is possible that it is difficult for the latent variable to learn the embedded features to distinguish different orientations of transparent objects. Also, it is difficult for the first stage detector to provide accurate location of the transparent objects, which heavily influences the second stage translation and orientation estimation. Since PMCL is provided with ground truth labels and search space, it performs comparatively well in the testset. However, PMCL uses single-view DLV as matching target which includes noise from specularity and distortion from transparent surfaces. Furthermore, DLV construction is computationally expensive, which takes an average 300 seconds for one object. In conclusion, \lite pipeline provides better accuracy than all three baseline methods on the testing dataset with a relatively small computationally cost. 

% More qualitative results of \lite are shown in Figure~\ref{fig:qual}.

\begin{table}[h]
    \centering
    \footnotesize
    \begin{tabular}{c|c|c|c|c|c|c|c}
    \hline
    AUC & wc & tc & gj & cc & sb & all & time(s)/obj\\
    
    \hline\hline
        DOPE & 0.14 & 0.16 & 0.21 & 0.16 & 0.00 & 0.18 & $<$ 1\\
        \hline
        AAE & 0.04 & 0.15 & 0.10 & 0.05 & 0.32 & 0.08 & $<$ 1\\
    \hline
        PMCL & 0.24 & \textbf{0.32} & 0.46 & 0.28 & 0.34 & 0.32 & 300\\
    \hline
        \textbf{\lite} & \textbf{0.38} & \textbf{0.32} & \textbf{0.62} & \textbf{0.35} & \textbf{0.44} & \textbf{0.45} & $<$ 10\\
    \hline
    \end{tabular}
    \caption{Comparison of \liteend, DOPE, AAE, and PMCL on transparent object pose estimation. The column headings wc, tc, gj, cc, and sb refer to the wine cup, tall cup, glass jar, champagne cup, and starbucks bottle objects, respectively. All columns, except for the last, refers to the area under the curve (AUC) for accuracy-threshold values for the symmetric objects metric (ADD-S), shown in Figure~\ref{fig:pose_comp}.}
    \label{tab:pose_comp}
\end{table}

\begin{figure}
    \centering
    \includegraphics[width=\columnwidth]{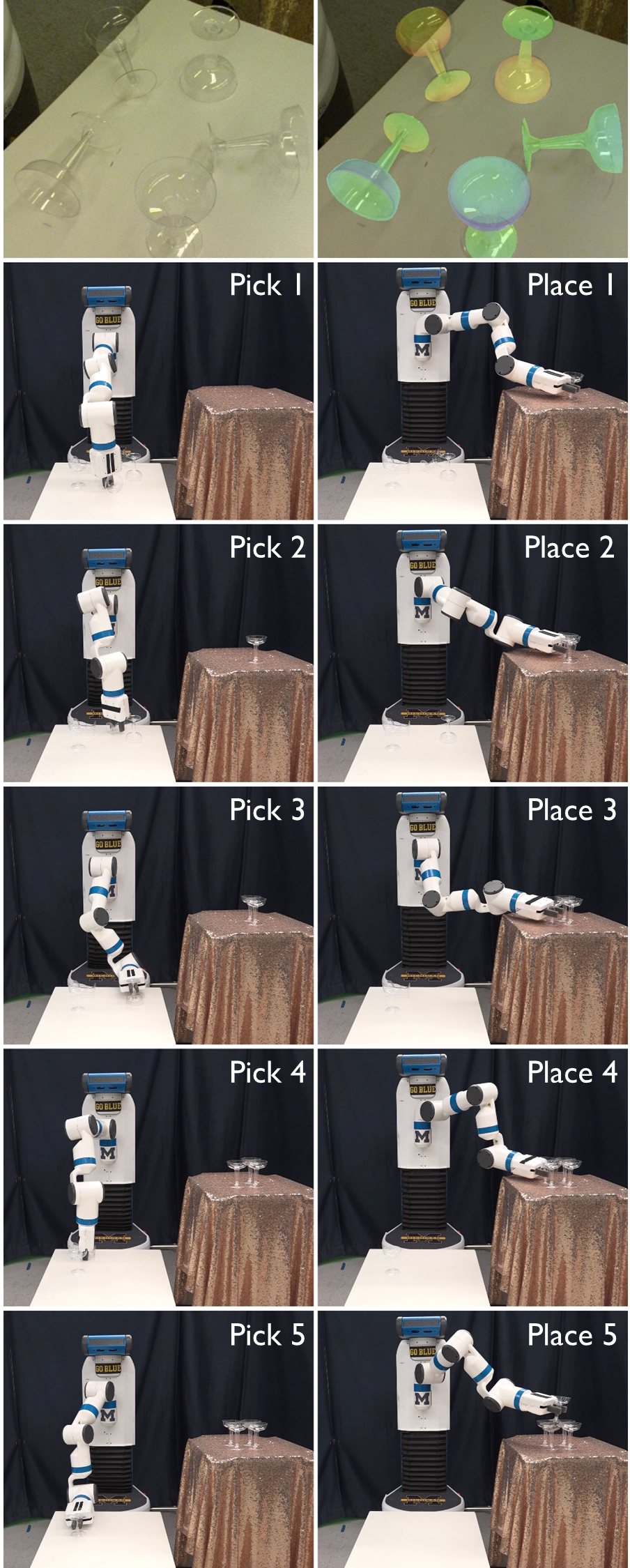}
    \caption{The robot is building a champagne tower by successfully picking and placing champagne cups on the table. The first row shows light-field observation (left) and pose estimation result from \lite (right). The following five rows show pick and place actions to finish the champagne tower.}
    \label{fig:robot_demo}
\end{figure}

\subsection{Champagne Tower Demonstration}
\lite is also integrated into a robotic manipulation pipeline for a purposeful manipulation task of building a champagne tower in a sparsely textured environment, as shown in Figure~\ref{fig:robot_demo}. In the initial setup, the champagne cups are randomly placed on a textureless white table. The Lytro Illum camera takes a light-field image and transfer the image with on-chip wifi. The Lytro camera's extrinsic matrix is calibrated with robot world frame. \lite then performs pose estimation over the scene, and the results are then adopted to transform the pre-defined grasp poses from the object's local coordinate frame to the robot world frame. With the accurate pose estimates, the robot is able to pick up all champagne cups from the table and arrange them into a champagne tower.

\section{CONCLUSIONS}
We introduce \liteend, a two-stage generative-discriminative object and pose recognition method for transparent objects using light-field observations. \lite employs the learning power of deep networks to distinguish transparent objects across light-field sub-aperture images. We show that the network trained only on synthetic data can deliver a good segmentation on transparent materials, which is served as matching target for second stage pose estimation. 
Along with the method, we propose the light-field transparent object dataset including synthetic and real data for the tasks of object recognition, segmentation, and 6D pose estimation. We demonstrate the use of \lite for a purposeful robot manipulation task over transparent cups. However, our method still has limitations in cluttered environments where the first stage segmentation results cannot provide distinguishable object shapes for second stage refinement.
Possible future works built on \lite could be instance-level segmentation based on transparent objects and single-view light-field depth estimation directly predicted by neural network.

\balance

                                  % on the last page of the docume+nt manually. It shortens

%%%%%%%%%%%%%%%%%%%%%%%%%%%%%%%%%%%%%%%%%%%%%%%%%%%%%%%%%%%%%%%%%%%%%%%%%%%%%%%%
%\section*{ACKNOWLEDGMENTS}

%This work was supported in part by Office of Naval Research grant N00014-08-1-0910 and NASA grant NNX13AN07A.

%%%%%%%%%%%%%%%%%%%%%%%%%%%%%%%%%%%%%%%%%%%%%%%%%%%%%%%%%%%%%%%%%%%%%%%%%%%%%%%%

\bibliographystyle{unsrt}
\bibliography{ref}

\end{document}